\documentclass[10pt,twocolumn,letterpaper]{article}
\pdfoutput=1
\usepackage{iccv}
\usepackage{times}
\usepackage{epsfig}
\usepackage{graphicx}
\usepackage{amsmath}
\usepackage{amssymb}
\usepackage{caption}
\usepackage{verbatim}

\usepackage[breaklinks=true,bookmarks=false]{hyperref}

\iccvfinalcopy % *** Uncomment this line for the final submission

\def\iccvPaperID{2551} % *** Enter the ICCV Paper ID here

\ificcvfinal\fi

\begin{document}

%%%%%%%%% TITLE
\title{3D-TalkEmo: Learning to Synthesize 3D Emotional Talking Head}

\author{Qianyun Wang \and Zhenfeng Fan \and Shihong Xia \\
Institute of Computing Technology, Chinese Academy of Sciences, Beijing, China
}
\twocolumn[{
\renewcommand\twocolumn[1][]{#1}
\maketitle
\begin{center}
    \centering
\includegraphics[width=0.9\linewidth]{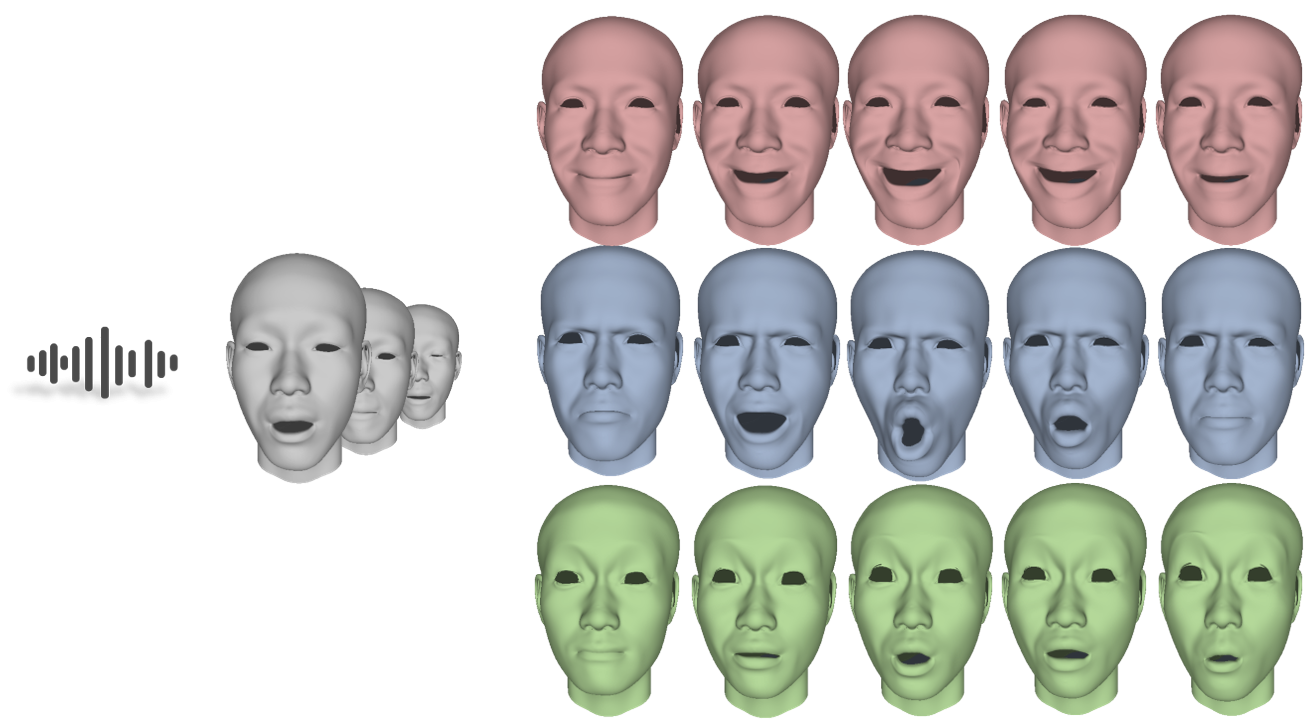}
    \captionof{figure}{3D-TalkEmo is able to animate a 3D talking head with various emotions. Please also see our \textbf{supplementary video}.}
    \label{fig:start}
\end{center}
}]
\ificcvfinal\thispagestyle{empty}\fi

%%%%%%%%% ABSTRACT
\begin{abstract}
   Impressive progress has been made in audio-driven 3D facial animation recently, but synthesizing 3D talking-head with rich emotion is still unsolved. This is due to the lack of 3D generative models and available 3D emotional dataset with synchronized audios. To address this, we introduce 3D-TalkEmo, a deep neural network that generates 3D talking head animation with various emotions. We also create a large 3D dataset with synchronized audios and videos, rich corpus, as well as various emotion states of different persons with the sophisticated 3D face reconstruction methods. In the emotion generation network, we propose a novel 3D face representation structure - geometry map by classical multi-dimensional scaling analysis. It maps the coordinates of vertices on a 3D face to a canonical image plane, while preserving the vertex-to-vertex geodesic distance metric in a least-square sense. This maintains the adjacency relationship of each vertex and holds the effective convolutional structure for the 3D facial surface. Taking a neutral 3D mesh and a speech signal as inputs, the 3D-TalkEmo is able to generate vivid facial animations. Moreover, it provides access to change the emotion state of the animated speaker. 
   We present extensive quantitative and qualitative evaluation of our method, in addition to user studies, demonstrating the generated talking-heads of significantly higher quality compared to previous state-of-the-art methods.
\end{abstract}

%%%%%%%%% BODY TEXT
\section{Introduction}
Audio-driven facial animation has gained increasing attention in recent years, which takes audio as input and face movement as output. The animated face can translate the content in audios to the face in videos, mostly manifesting as mouth movement. These facial animation technologies can be used in varies field such as virtual conference, film-making, and computer games.

\par Audio-driven face animation methods can be divided into 2D and 3D methods by the data they act on. In the 2D case, animation of facial image has undergone tremendous progresses with the development of convolutional neural networks (CNNs). The animated faces are easily transferable with different expression, hair style, skin color, etc. In the 3D case, the researchers are now paying more and more attention to the animation of point cloud data, due to the advances in 3D image sensors and the potential applications with the animated faces.  
\par However, while there are a lot of studies ~\cite{Taylor17deep, CudeiroBLRB19, DBLP:journals/tog/KarrasALHL17, DBLP:journals/tog/ZhouXLKMS18} for audio-driven 3D facial animation, the emotion state of animated faces are commonly neglected in the literature. Since emotion is a born character of real humans, animation of faces without emotion is like a doll losing the soul. Rendering emotion on an animated face can also raise the interest of audience by strong visual impact.
While the previous studies on audio-driven facial animation focus on the mouth movement to synthesize a speaker, the emotion state is rarely taken as an instructive input. Emotion is defined as the micro-expression of a human, which manifests as some subtle movements of muscles on a face. 

\par Since mapping between audio and face movement is a many-to-many problem, it poses great challenge to synthesize face across identities and speaking styles. Among them, 3D modeling of realistic emotions remains an unsolved problem for several reasons. Firstly, one can take the deep learning methods as powerful fitting machines, but there is a shortage of data for 3D scans of faces with both synchronized audios and emotion labels. Secondly, the existing network architectures are mostly for 2D images and not applicable to 3D geometric data. Finally, while most existing works focus on the action of movement around the mouth regions in the audio-driven facial animation literature, modeling of emotions that correlate to small movements of muscles on the whole facial region is particularly difficult.

\par This work proposes an audio-driven 3D facial animation method for unpaired emotion transfer. We refer to the proposed model as ``\textbf{3D-TalkEmo}'' (see Figure~\ref{fig:start}). We mean by ``unpaired'' a model that is trained with data without exact correspondences for corpus of audio and different emotion states, which takes into consideration the fact that exactly paired data are almost unachievable. 
We propose a basic framework that allows one to reconstruct a dynamic 3D facial mesh driven by an audio sequence. It also provides flexibility for users to modify the emotion of the output speakers as 3D data. We achieve this by the recent sophisticated technologies of 3D face reconstruction, phonetic features extraction, and unpaired style transferring. In summary, our main contributions are as follows:
\begin{itemize}
\item[-] We create a large 3D dataset with synchronized audios and videos, rich corpus, as well as various emotion states of different persons with the sophisticated 3D face reconstruction method and 3D face models.
\item[-] We propose an end-to-end framework for 3D talking head animation, which takes a static 3D facial mesh and a period of audios as input, and outputs a 3D sequence of talking head with various emotions.
\item[-] We address the problem of emotion transfer on 3D facial meshes using a multi-dimensional scaling based projection method, which preserves the local neighborhood relationship of each vertex in a least-square sense. It is the first method providing access for unpaired emotion transferring on 3D facial meshes with deep learning to the best of our knowledge. 
\end{itemize}

%------------------------------------------------------------------------
\section{Related Work}
What are closely related to our work include the existing literature in audio-driven facial animation and emotion modeling. Both of the two fields have received increasing attentions recently. We now give brief reviews of them by covering the most important works, respectively.

\begin{figure*}
\begin{center}
\includegraphics[width=1.0\linewidth]{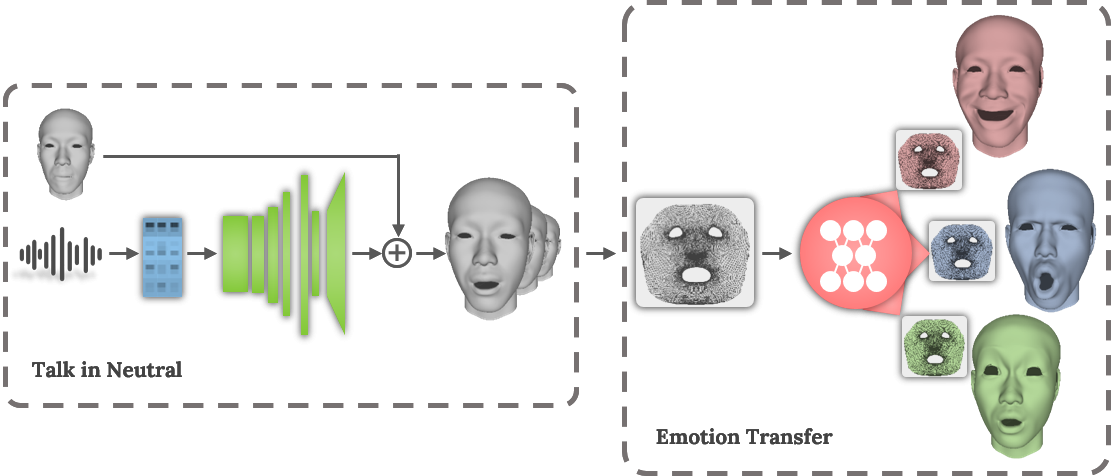}
\end{center}
   \caption{The pipeline of 3D-TalkEmo.}
\label{fig:pipeline}
\end{figure*}

%------------------------------------------------------------------------
\subsection{Audio-driven Facial Animation}
Many existing works attempt to drive a cartoon or a true face using synchronized audios. The output is usually a segment of realistic videos composed of 2D facial images or 3D meshes of a speaker. We briefly discuss the methods and the training dataset as follows.

\par Taylor \emph{et al.}~\cite{Taylor17deep} use recorded videos of 8 hours as training data and divides the input phonetic transcript to 8 basic categories. Then they employ a deep CNN with sliding windows to learning the mapping between the phonemes and the coefficients from the active appearance model (AAM)~\cite{cootes2001active}. The AAM coefficients are finally used to render a facial rig for animation.

\par Inspired by the above 2D method, Karras~\emph{et al.}~\cite{DBLP:journals/tog/KarrasALHL17} directly learn the mapping from normalized audio waveform to 3D vertices with a sliding window and a CNN architecture. Their dataset include a series of recorded videos ranging from 3 to 5 minutes. They also learn the emotion representation with some manual labels, but only for $2$ persons. 

\par Suwajanakorn \emph{et al.}~\cite{Suwajanakorn17synthesizing} employ a LSTM~\cite{HochreiterS97} structure to learn the mapping from Mel-frequency cepstral coefficients (MFCCs) to the coefficients of an active shape model (AAM)~\cite{cootes1995active}. Their training data include $17$ hours' 2D talking videos of President Obama. The animated output includes realistic lip movement and are well matched to various audios.

\par Zhou \emph{et al.}~\cite{DBLP:journals/tog/ZhouXLKMS18} propose a Visemenet model to animate a facial rig named JALI~\cite{DBLP:journals/tog/EdwardsLFS16}. Their model includes a LSTM architecture assisted by the facial action unit system (FACS)~\cite{Ekman1978FacialAC} and viseme motion curve (VMC)~\cite{Fisher1968ConfusionsAV}. The dataset include $15$ hours of paired audio and video data. Various phoneme groups requires to be labelled manually. 

\par Cudeiro \emph{et al.}~\cite{CudeiroBLRB19} propose a CNN based model, which is namely VOCA, to learn the mapping of audio to the residual offset of 3D vertices for a face model. DeepSpeech~\cite{HannunCCCDEPSSCN14} is used as a robust feature extractor for the input audio signals. This model is trained with recorded 4D scans of 29 minutes with several different persons. 

\par Zhou \emph{et al.}~\cite{DBLP:journals/corr/abs-2004-12992} propose a method that generates photorealistic videos of entire talking heads with various 2D characters with audio as the only input, which disentangles the content and speaker information in the input audio signal.

\par The above existing methods on audio-driven facial animation have at least one of the following limitations.
1) The animation is limited to 2D images or a few exemplar 3D faces;
2) The methods requires extensive efforts on manually handling;
3) The methods focus on movements of the mouth regions regardless of the emotion state that correlates to the whole facial region.

%------------------------------------------------------------------------
\subsection{Emotion Modeling}
\par The existing works model the emotion state of a talking face either discretely or continuously as a vectors of several dimensions. The specific components include happiness, sadness, anger, disgust, fear, surprise, \textit{etc.} Since the related deep learning based methods rarely discuss the emotion modeling, we only discuss several representative traditional methods.

\par Mehrabian and Russell~\cite{mehrabian1974approach} propose the PAD (pleasure-displeasure, arousal-non-arousal, dominance-submissiveness)
emotion model, which include different degrees of certain emotions. This model also considers that emotions have three different dimensions of pleasure, activation, and dominance.

\par Whissell \emph{et al.}~\cite{whissell2009using} propose a two-dimensional emotion model with arousal-valence-control, where arousal indicates the arousal of emotion and valence is the degree of positivity. Both dimensions are represented by numerical values. It is demonstrate that almost all emotions of human faces can be included in this two-dimensional space.

\par Cao \emph{et al.}~\cite{cao2005expressive} decompose facial movements into speech part and emotion part via Independent Component Analysis (ICA). They also uses a radial basis function (RBF) to transform the emotion space to a canonical space with smaller dimensions.

\par Zhang \emph{et al.}~\cite{jia2010emotional} build an emotional text-to-audio-visual speech (ETTAVS)
system based on a 3D emotion model, and takes as input
a 3-dimensional PAD model~\cite{mehrabian1996pleasure} to calculate partial expression parameters (PEP)~\cite{wu2006real}. The PEP describes the local facial motion. Finally, the output is some facial animation parameters (FAPs)~\cite{wu2006real, doi:https://doi.org/10.1002/0470854626.ch9} which is capable of converting neutral emotions to other different emotions. 

In a departure from the traditional method that models the emotions explicitly, this paper models the emotion implicitly as simple discrete labels. We expect that the proposed CNN architecture can learn these emotions automatically with tremendous training data.

%------------------------------------------------------------------------

\section{Method}
The pipeline of our method is illustrated in Figure~\ref{fig:pipeline}. Automatically generating emotional 3D face animation from input audio involves three steps: 1) Extract speech feature of the input audio; 2) Predict vertex displacement for 3D face from the input speech feature; 3) Emotion transfer and generation of 3D mesh data. We borrow the effective network architecture from the state-of-the-art DeepSpeech model~\cite{DBLP:journals/corr/HannunCCCDEPSSCN14} and VOCA model~\cite{CudeiroBLRB19} for the first two steps. On this basis, we create a 3D dataset with various identities and corpus, and different emotion states using the state-of-the-art 3D face reconstruction methods, and propose to a novel network architecture to address the unpaired emotion transfer problem on 3D meshes.

\begin{figure}[t]
\begin{center}
\includegraphics[width=1.0\linewidth]{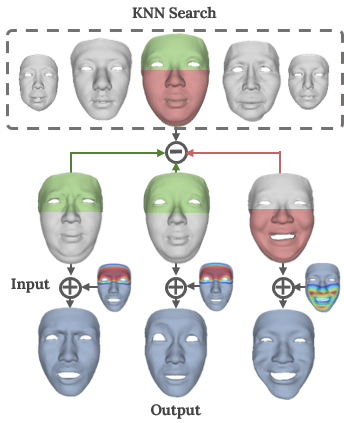}
\end{center}
   \caption{The pipeline for emotion augmentation.}
\label{fig:Emotion Augmentation}
\end{figure}

\begin{figure*}
\begin{center}
\includegraphics[width=1.0\linewidth]{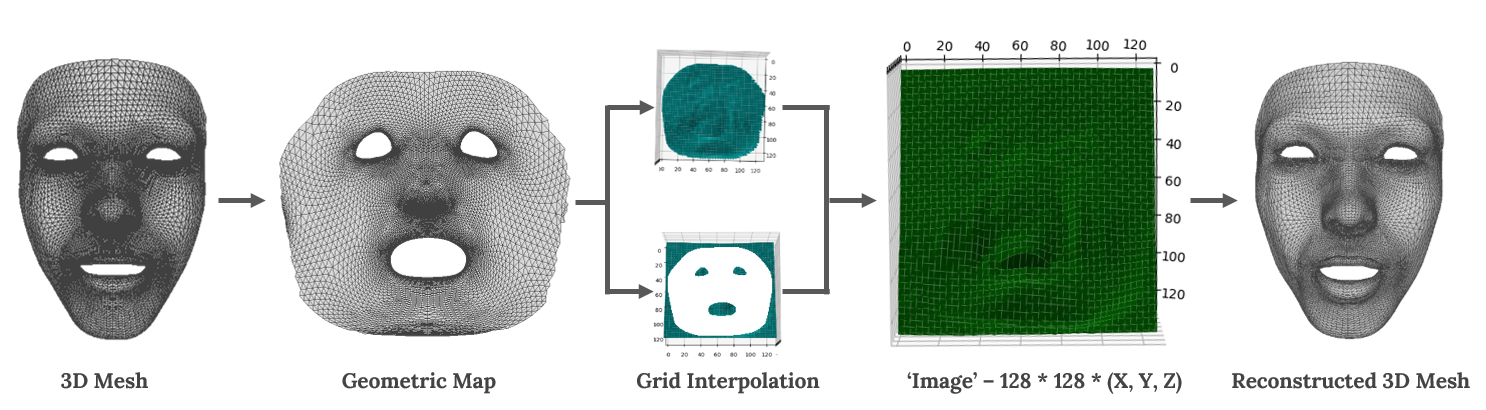}
\end{center}
   \caption{The process of mapping a 3D facial surface to the canonical geometric map using MDS.}
\label{fig:MDS}
\end{figure*}

\subsection{Construction of 3D Emotional Talking Heads}
\par Talking face animation is unachievable without audio-synchronized data with variations on both corpus and emotion. However, most existing dataset with synchronized audios are not enough to cover the corpus and emotion fully. The dataset with rich emotion (\textit{e.g.}~~\cite{livingstone2018ryerson}) is usually limited by the amount of corpus, while the speaker's emotion in other dataset with sufficient corpus (\textit{e.g.}~~\cite{Suwajanakorn17synthesizing}) is mostly neutral. Besides, unlike 2D images, emotional performance on 3D data format are always more expensive, which usually require professional actors, expensive devices for data collection, and complex data processing process. Even the real humans may have the problem of insufficient emotion. Therefore, it is necessary to create a 3D face dataset with various identities and rich emotion at low cost while maintaining the amount of corpus.

\par To the best of our knowledge, there is no such large-scale 3D dataset in the literature. To overcome this difficulty, our key idea is to reconstruct 3D face from input images, then we augment the emotion of the mesh to be more realistic and expressive.

\par \textbf{3D face reconstruction.} We reconstruct the 3D facial mesh from the input videos using the 3DMM method~\cite{blanz1999morphable} with a recent published FaceScape model~\cite{yang2020facescape}. The FaceScape model is capable of modeling both the identity and expression variations as  
\begin{equation}\label{biliner model}
V = C_r \times w_{exp} \times w_{id},
\end{equation}
where $C_r$ is a 3-dimensional core tensor as the 2D principal components, $w_{exp}$ is the expression parameters, and $w_id$ is the identity parameters. 
This is a bilinear model for the representation of a facial shape. Then, the reconstruction process is to minimize the following function:
\begin{equation}\label{3DReconstruction}
E(\beta ,s,R,T) = \sum\limits_i {{{\left\| {{x_i} - SOP(R,T,s,{V})} \right\|}^2}},
\end{equation}
where $x_i$ is the location for each pixel and $SOP(\cdot)$ denotes scaled orthographic projection with respect to the rotation $R$, translation $T$, scaling factor $s$, and 3D shape $S_\beta$ as
\begin{equation}\label{3D2DProjection}
SOP(R,t,s,{V}) = s\left[ {\begin{array}{*{20}{c}}
	{\begin{array}{*{20}{c}}
		1 & 0 & 0  \\
		\end{array}}  \\
	{\begin{array}{*{20}{c}}
		0 & 1 & 0  \\
		\end{array}}  \\
	\end{array}} \right]R{V} + T.
\end{equation}

We only use the landmarks instead of all pixels for the optimization of Eq.~\ref{3DReconstruction} for computational efficiency. The 2D landmark location can be estimated by rendering of 3D landmarks using Eq.~\ref{3D2DProjection}. Optimizing Eq.~\ref{3DReconstruction} and Eq.~\ref{3D2DProjection} alternatively will lead to the results for 3D reconstructions. In practise, we fix the identity parameter $w_{id}$ after the reconstruction via several frames at the beginning of a facial video.

We apply the above method to around $10,000$ audio-synchronized videos from the RAVDESS dataset\cite{livingstone2018ryerson} and the LBG dataset\cite{alghamdi2018corpus}, assisted by a state-of-the-art facial landmark detector~\cite{wang2019adaptive}. 

\par \textbf{Emotion augmentation.} We use four basic and easily distinguishable emotions in our method: \textit{neutral, happy, angry and surprise}. Because the emotion state of reconstructed 3D face using only detected landmarks is not prominent enough, we further enhance the data for the non-neutral emotions. By a careful checking of the real 3D data in the FacesScape dataset, we find that angry and surprise are mostly represented as vertex displacements near the eyebrows (upper face), while happy is related to the displacements of vertex near the cheekbone (lower face). Therefore, we augmented the emotion of each reconstructed mesh in three steps: Firstly, we select the data with calm, eyebrows up, eyebrows down, and grin expressions in the FaseScape dataset as reference candidate; Then, we calculate the weight of the upper/lower face drawn by an animator, which is 
proportional to certain boundary (\textit{e.g.} the inner edges for the eyes or the mouth); Finally, we use nearest neighbor search algorithm to find the 3D model with calm expression in the FaceScape dataset whose feature point is closest to the reconstruct mesh, and increase the displacement based on the weighted displacement given by the error between the emotional model and the calm model given by nearest neighbor search. The pipeline is illustrated in Figure~\ref{fig:Emotion Augmentation}.

%------------------------------------------------------------------------

\par Since the selected 2D video data on the RAVDESS dataset \cite{livingstone2018ryerson} and the LBG dataset \cite{alghamdi2018corpus} are with synchronized audios, an audio-synchronized 3D face data set is constructed accordingly. Therefore, the mapping from the input voice to the displacement of vertices on a 3D face model can be learned with this constructed dataset. 

%After the construction of the voice-synchronized 3D face data set is completed, the mapping from the input voice to the 3D face model can be learned. On the one hand, the noise, accent, and language of the speech may affect the final animation; on the other hand, due to the different speaking styles of the performers, such as Obama’s tendency to speak. Even if all speakers are in neutral emotions, the mapping from emotion to the vertex displacement is a challenging one-to-many problem. Therefore, we propose to extract speech feature robustly and learn different speaking style of 3D models with the reconstructed dataset including many different identities.

\subsection{Representation of 3D Mesh as Image}\label{Sec:MDS}
\par A straightforward way to manipulate a facial mesh is to directly use a neural network for predicting coordinates of each mesh vertex, like PointNet\cite{DBLP:conf/cvpr/QiSMG17}. However, the emotion of a facial mesh is usually represented as small displacements compared to the mouth movement during speaking. Predicting the coordinates directly neglects the adjacency information and is prone to distortion of the resulted mesh. 
%A failure case is shown in Figure \ref{fig:Qualitative Results}. 
On the other hand, the existing 3D network architectures such as PointNet and PointNet++~\cite{10.5555/3295222.3295263} are demonstrate to work well for a number of tasks including point cloud recognition and segmentation. However, they are not suitable for model the emotions as subtle movement of faces, either. 
They do not work well for the emotion transfer task on a facial mesh, which is a piecewise smooth surface.
Therefore it is necessary to develop a new method that can take effective use of the adjacency information for emotion transfer on the 3D facial surface.

\par In this study, we proposed a novel method that translates the adjacency relationship of a facial surface to a canonical 2D images. We denote the generated 2D image as the geometric map that encodes the shape of a facial mesh. The 2D geometric map provides access to utilize the advanced technologies on 2D facial images for domain transfer. The construction for the 2D canonical geometric map is described as follows:

(1) Calculate all the vertex-to-vertex geodesic distance on a template 3D facial mesh;

(2) Apply classical multi-dimensional scaling (\textbf{MDS})~\cite{cox2008multidimensional} analysis to the distance matrix, and keep the first $2$ dimensional of the result;

(3) Map the resulted $2$-dimensional coordinate to 2D images, and perform grid sampling with respect to each 3D coordinate to re-order the geometric map to a regular grid within a mask.

\par Figure \ref{fig:MDS} illustrates the process. We adopt a fast heatmap diffusion method~\cite{DBLP:journals/tog/CraneWW13} for the computation of the vertex-to-vertex geodesic distances. We also use an ensemble of all samples in our constructed dataset as the template face. \emph{It is worth mentioning that the classical MDS preserves the distance metric (geodesic in this case) in a least-square sense, thus preserving the adjacency information on the 3D facial surface in an optimal way}. By this way, we represent a 3D facial mesh $(V,E)$, where $V$ denotes the set of vertices and $E$ denotes the set of edges, to a 2D image with the size $128\times128\times3$, where the 3D coordinates is stored in the third dimension within a mask. We refer to the above representation as geometry map and use it in our emotion transfer network.

%------------------------------------------------------------------------

\subsection{Neutral Talking Model}

\par We use DeepSpeech~\cite{DBLP:journals/corr/HannunCCCDEPSSCN14} to extract speech features and borrow the network architecture from VOCA~\cite{CudeiroBLRB19} for our neutral talking model. Given an input audio of $T$ seconds with 60 frames per second(fps), the DeepSpeech model output a feature of size $30T \times D$, where $D$ is the number of characters in the alphabet plus one for a blank label. We upsample the output in the time axis to match the target video using linear interpolation. In order to incorporate temporal information, we convert the audio frames for overlapping windows of size W, thus the output is further a three dimensional array $60T \times W \times D$.
The output of DeepSpeech model is then fed into the input for the VOCA model.

\begin{figure}[t]
\begin{center}
\includegraphics[width=0.6\linewidth]{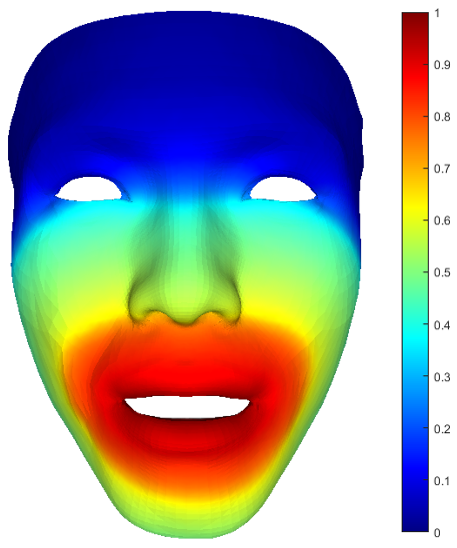}
\end{center}
   \caption{The weights to adjust the loss function for the VOCA model.}
\label{fig:Local Weight}
\end{figure}

For the second stage of the encoder-decoder network, we keep the basic setting as the original VOCA model. In a departure from the original work, the 3D data we use have the resolution of $12,483$ for the facial region, which is higher than than $5,023$ reported in~\cite{CudeiroBLRB19}.Moreover, we adjust the position term in the loss function by a weight (see Figure~\ref{fig:Local Weight}) that focuses its attention on the mouth regions, which benefit the final reconstruction results.

We train the VOCA model with the neutral samples in our reconstructed dataset.

\subsection{Emotion Transfer}

\par Based on a 3D facial video dataset with emotions for a variety of identities and corpus, the emotion transfer is considered as an unpaired domain transfer problem in this work. While domain transfer and adaption are active research areas on 2D images in the literature, its applicability to 3D data remains an unsolved problem. This is due to difficulties for mining the adjacency information of 3D data.  

We propose to use the geometric map generated by MDS as described in Sec.~\ref{Sec:MDS}. The geometric map encodes the 3D coordinates as 2D images while preserving the adjacency information in a least-square sense. This provides access for us to incorporate the state-of-the-art methods in 2D images, and we use the StarGAN~\cite{choi2018stargan} as the basic network architecture, which is a network for multi-domain image-to-image translation, for our emotion transfer task. Similar to the StarGAN, we adopt the following loss functions for training:

\par \textbf{Adversarial loss.} The adversarial loss forces the distribution of the generated images to approach the real images, as
\begin{equation}\label{biliner model}
L_{adv} = E_x[\log D_{src}(x)] + E_{x,c}[\log (1-D_{src}((G(x,c))],
\end{equation}
given the generator $G(x,c)$ conditioned on both the input image $x$ and the target domain label $c$, while $D$ is a discriminator.

\par \textbf{Domain classification loss.} To translate $x$ into an output image $y$, which is properly classified to the target domain $c$. the objective is decomposed into two terms: a domain classification loss as
\begin{equation}\label{ablation-local}
L_{cls}^r = E_{x,c}[-\log D_{cls}(c'| x)]
\end{equation}
for real images to optimize $D$, and a domain classification loss 
\begin{equation}\label{biliner model}
L_{cls}^f = E_{x,c}[-\log D_{cls}(c|G(x,c))]
\end{equation}
for fake images to optimize $G$.

\par \textbf{Reconstruction loss.} A cycle consistency loss~\cite{zhu2017unpaired} is employed as
\begin{equation}\label{ablation-augmentation}
L_{rec} = E_{x,c,c'}	\left \| x-G(G(x,c),c')_1 \right \|
\end{equation}
for the generator to preserve the content of its input images while changing only the domain-related part of the inputs.

\par \textbf{Full objective} The objective functions to optimize $G$ and $D$ are combination of the above loss, respectively, as 
\begin{equation}\label{biliner model}
L_D = -L_{adv} + \lambda_{cls} L^r_{cls}
\end{equation}
and
\begin{equation}\label{biliner model}
L_G = L_{adv} + \lambda_{cls}L^f_{cls} + \lambda_{rec}L_{rec},
\end{equation}
 where $\lambda_{cls}$ and $\lambda_{rec}$ are hyper-parameters that balance the domain classification and reconstruction errors, respectively.

%------------------------------------------------------------------------
\section{Experiments}

\subsection{Implementation Details}

We implement the proposed method on the PyTorch platform. On the first stage for training the neutral talking model, we use the Adam optimizer and the initial learning rate is set to $1\times e^{-4}$. The mini-batch size is set to $64$ and number of epoch is $100$. On the second stage for the emotion transfer network, we set the number of iteration to $200,000$. Other settings follow the StarGAN network~\cite{choi2018stargan}. It takes around 5 hours to train the whole network with our created 3D video dataset with a single GPU (GTX2080).

%------------------------------------------------------------------------
\subsection{Latent Space Visualization}
\par
\begin{figure}[t]
\begin{center}
\includegraphics[width=1.0\linewidth]{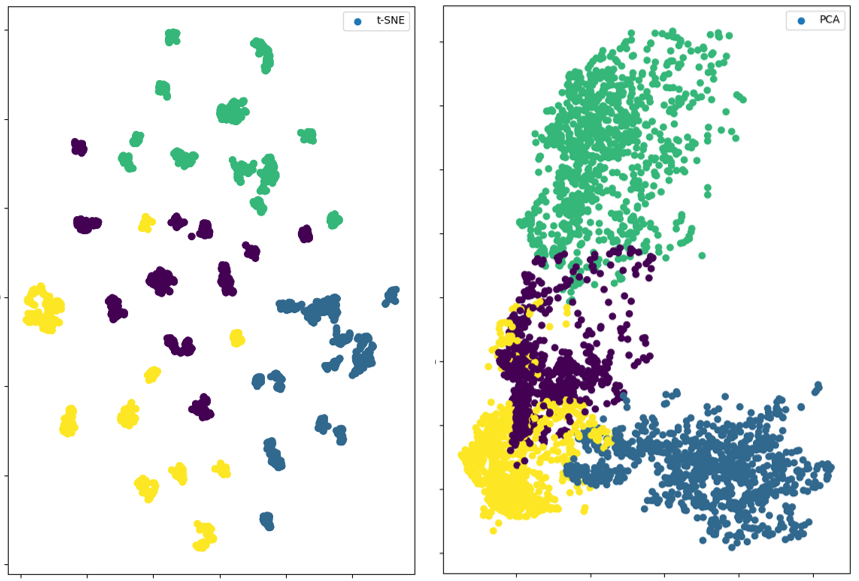}
\end{center}
   \caption{Latent space Visualization by the t-SNE (the left) and PCA (the right), respectively.}
\label{fig:Lantent space Visualization}
\end{figure}

 In this experiment, we project the output feature vectors at the end of the emotion transfer network onto a 2D coordinate system to view the clustering effect of the data. Principal component analysis (PCA) and t-distributed stochastic neighbor embedding (t-SNE) are two dimension reduction methods we employed to view the data. Figure~\ref{fig:Lantent space Visualization} shows the results for two methods, respectively. The $4$ different emotions of calm, angry, happy, and surprise are distributed reasonably on the 2D plane, which implies that the proposed emotion transfer network 
has learned the representation of different emotion states well. Note that the dispersion of data with the same emotion state by t-SNE is caused by different identities.

%by using t-distributed stochastic neighbor embedding (t-SNE) and , and plot the results in order to gain a better understanding of how the network interprets content and style in practice.
%Figure 5 shows the 2D projection of our style parameters, where each sample is marked with a color corresponding to its style label. It can be seen that our network learns to cluster the style parameters, which means that style inputs that share the same style will manipulate the motion content in a similar way. This result demonstrates that the extracted style parameters mostly depend on the style label.

%\begin{itemize}
%\item 2*2 code projected onto 2D space using t-SNE
%\item 2*2 code projected onto 2D space using PCA
%\end{itemize}

%------------------------------------------------------------------------

\subsection{Quantitative Evaluation}
\par \textbf{Evaluation metrics}. To evaluate the accuracy of animated faces for lip movements, we extract the corresponding landmarks around the mouth region and apply them to some quantitative evaluation metrics, as follows:
\begin{itemize}
\item Reconstruction error (\textbf{RE}): the average Euclidean distance between the coordinates of all predicted vertices and reference ones.
\item Velocity error (\textbf{VE}): the average Euclidean distance between the inter-frame distances of all predicted vertices and reference ones. This metric serves as an indicator for accurate dynamic vertex motion.
\item Classification error (\textbf{CE}): classification error of the expression of the animated facial meshes using the discriminator of the emotion transfer network.
\end{itemize}

\par \textbf{Test set split}.We created a test split from the reconstructed dataset by some samples of RAVDESS~\cite{livingstone2018ryerson} and LBG dataset~\cite{alghamdi2018corpus}, containing $500$ video segments from $27$ speakers. Each video segment lasts $2$ to $4$ seconds. These speakers' identities are not included in the training set.  

\par \textbf{Effect of local weight}. In this experiment we test the effect of the weighting strategy for the training of the neutral talking model. We set the weights to be a constant $1$ as in the original VOCA model for comparison. The results are reported in Table \ref{table:ablation-local}. We can see that the proposed weighting strategy leads to lower reconstruction and velocity errors, indicating realistic facial animation.
\par \textbf{Effect of emotion augmentation}. In order to show the effectiveness of the data augmentation strategy for the training of the emotion transfer network, we train our model on the reconstructed dataset without augmentation. The results for the generator and discriminator are shown in Table~\ref{table:ablation-augmentation}. We can see that the proposed emotion augmentation strategy leads to lower classification errors for discriminator network, indicating well discrimination for different emotions. The over-smoothing results of the original dataset also lead to smaller errors for the generator, however, the ability to model various emotions state is weakened.

\begin{table}
\begin{center}
\begin{tabular}{lcccc}
Method & T-RE & T-VE & V-RE & V-VE\\
\hline
VOCA & 1.573 & 1.226 & 7.728 & 1.116 \\
% Add Local Weight & \textbf{0.1159} & \textbf{0.08695} & \textbf{0.5505} & \textbf{0.1137}\\
Local Weight & \textbf{0.116} & \textbf{0.087} & \textbf{0.551} & \textbf{0.114}\\
\hline
\end{tabular}
\end{center}
\caption{Evaluation of the weighting strategy for the neutral talking model. The meaning of some symbols: T-RE: RE for training, T-VE: VE for training, V-RE: RE for validation, V-RE: VE for validation. Smaller is better.}
\label{table:ablation-local}
\end{table}

\begin{table}
\begin{center}
\begin{tabular}{lcccc}
Method & D-CE & D-RE & G-CE & G-RE\\
\hline
Original & 5.36 & \textbf{1.33} & \textbf{1.69} & 1.25\\
Augmentation & \textbf{2.09} & 1.58 & 7.43 & \textbf{0.95}\\
\hline
\end{tabular}
\end{center}
\caption{Evaluation of the emotion augmentation strategy for the emotion transfer model. The meaning of some symbols: D-CE: CE for the discriminator, D-RE: RE for the last feature layer of the discriminator, G-CE: CE for the generator, G-RE: RE for the generator. Lower is better.}
\label{table:ablation-augmentation}
\end{table}

%------------------------------------------------------------------------
\subsection{Qualitative Evaluation}

\par
\begin{figure}[t]
\begin{center}
\includegraphics[width=1.0\linewidth]{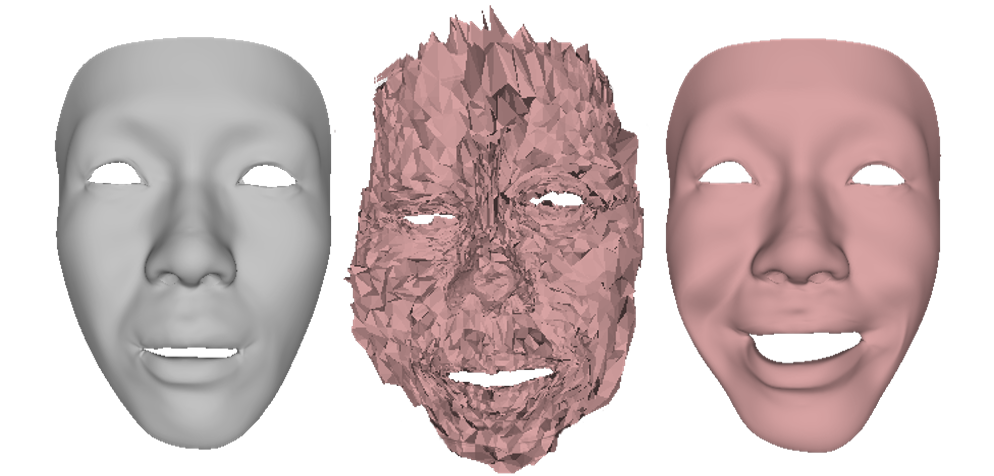}
\end{center}
   \caption{Qualitative comparison between the PointNet and the proposed geometric map with StarGAN as the generator networks for emotion transfer.}
\label{fig:Qualitative Results}
\end{figure}

 In this work, we propose to use the geometric map to project the 3D coordinate of a facial surface to the 2D plane, and then use the architectures of the StarGAN~\cite{choi2018stargan} for emotion transfer. Since there is no previous work for unpaired 3D emotion transfer, we use the PointNet~\cite{DBLP:conf/cvpr/QiSMG17} to replace our generator network for comparison. A qualitative results on a single frame of video are shown in Figure~\ref{fig:Qualitative Results}. We can see that the result by PointNet are contaminated by trenmendous noise, whereas our proposed method transfers the emotion well from calm to happy. We owe the success of the proposed method to the effective modeling of adjacency information between neighboring vertices on the facial surface.

\subsection{User Study}
\par Since the unpaired pipeline in this work is without the ground-truth, we conduct an additional user study experiment to validate the effectiveness of our method. We compare our full method with the one without local weight, the one without emotion augmentation, the one replaced by PointNet structure, and two state-of-the-art methods~\cite{CudeiroBLRB19,Karras17audio}.

Following the basic setting from~\cite{DBLP:journals/corr/abs-2004-12992}, each participant is shown a questionnaire with 15 queries involving random pairwise comparisons. For each query, we show $5$ generated videos by each of the methods above. $23$ different participants are asked with 3 types of questions, which we describe below.

\begin{itemize}
\item Realism: Participants are asked which character's facial expression look more realistic.
\item Content Preservation: Participants are asked which character speaks more similar to input audio.
\item Emotion Transfer: Participants are asked which character has richer emotion while speaking.
\end{itemize}

\begin{figure}[t]
\begin{center}
\includegraphics[width=1.0\linewidth]{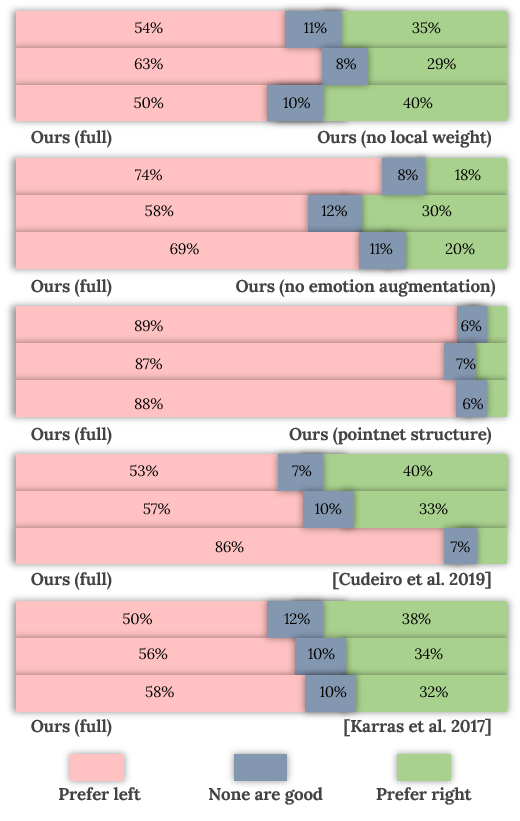}
\end{center}
   \caption{User study results for realism (top row in each group), content preservation (Second row in each group), and emotion transfer (bottom row in each group).}
\label{fig:User Study}
\end{figure}

Participants are asked to pick one of three choices ("left", "right", "none"). 345 responses are collected for this user study. Figure~\ref{fig:User Study} shows the study result. It can be seen that our method yields results which are more faithful to the task of synthesizing 3D talking head with rich emotion.  

%------------------------------------------------------------------------
\section{Conclusion}
In this paper, we propose 3D-TalkEmo, a novel audio driven facial animation model based on deep neural networks. Given a 3D facial mesh and a period of audio, the proposed model is capable of synthesize a realistic 3D talking head. It also allows one to customize the emotion state of the speaker with a emotion transfer network. The key element of our proposed emotion transfer network is the canonical geometric map by the classical MDS that maps the 3D representation of facial surface to the 2D planes. Extensive experiments demonstrate the effectiveness of the proposed method. We hope this work including the creation of 3D video dataset, will be helpful for future research.

%that transforms [...] into a 3D mesh, which is an extremely cross-domain task. To accomplish this task, we build a dataset and establish a PCA model, and propose two novel loss terms based on previous psychological study. Our method is fast and thus makes futher interaction control possible. We also propose a simple and effective method that allows a user to interactively adjust the results. Experiments and a user study demonstrate the effectiveness of our method.

{\small
\bibliographystyle{ieee_fullname}
\bibliography{egbib}
}

\end{document}